\documentclass[conference]{IEEEtran}
\usepackage{times}

\usepackage[numbers]{natbib}
\usepackage{multicol}
\usepackage[bookmarks=true]{hyperref}
\usepackage{times}
\usepackage{epsfig}
\usepackage{graphicx}
\usepackage{amsmath}
\usepackage{amssymb}
\usepackage{siunitx}
\usepackage{booktabs}
\usepackage{bm}
\usepackage{xcolor}

\begin{document}

\title{Robustness Meets Deep Learning: An End-to-End Hybrid Pipeline for Unsupervised Learning of Egomotion}




%
\author{\authorblockN{Alex Zihao Zhu\authorrefmark{1},
Wenxin Liu\authorrefmark{1},
Ziyun Wang, 
Vijay Kumar and
Kostas Daniilidis}
\authorblockA{University of Pennsylvania\\ Email: \{alexzhu, wenxinl, ziyunw, kumar, kostas\}@seas.upenn.edu}
\authorblockA{\authorrefmark{1}These authors contributed equally to this work.}}

\maketitle

\begin{abstract}
In this work, we propose a method that combines unsupervised deep learning predictions for optical flow and monocular disparity with a model based optimization procedure for instantaneous camera pose. Given the flow and disparity predictions from the network, we apply a RANSAC outlier rejection scheme to find an inlier set of flows and disparities, which we use to solve for the relative camera pose in a least squares fashion. We show that this pipeline is fully differentiable, allowing us to combine the pose with the network outputs as an additional unsupervised training loss to further refine the predicted flows and disparities. This method not only allows us to directly regress relative pose from the network outputs, but also automatically segments away pixels that do not fit the rigid scene assumptions that many unsupervised structure from motion methods apply, such as on independently moving objects. We evaluate our method on the KITTI dataset, and demonstrate state of the art results, even in the presence of challenging independently moving objects.
\end{abstract}

\IEEEpeerreviewmaketitle

\begin{figure*}
\centering
\includegraphics[width=0.9\linewidth]{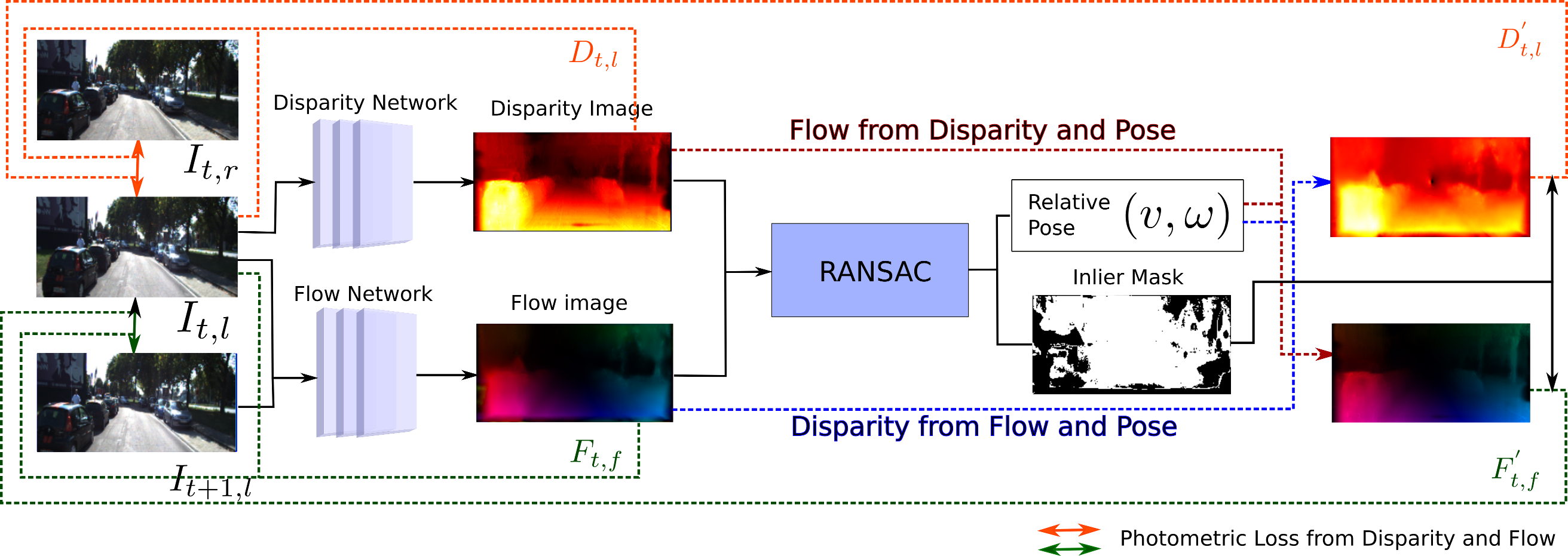}
\caption{Our pipeline accepts three images as input: the left and right images from a stereo pair, and the left image from the next time step. Dense disparity and optical flow maps are estimated from the stereo and temporal pairs, respectively, and passed into our RANSAC estimator to produce a robust estimate of the camera velocity, $v$, $\omega$, and an inlier mask. The estimated velocity is then used to estimate optical flow from disparity, and vice versa, using a rigid motion model, which allows us to apply an additional set of refinement photometric losses on the network flow, $F'_{t,f}$, and disparity outputs, $D'_{t,l}$, masked with the RANSAC inlier mask. Best viewed in color.}
\label{fig:pipeline}
\end{figure*}
\section{Introduction}
The advent of unsupervised methods for neural networks has resulted in the rapid growth of works that train networks that predict the camera pose and depth of a scene, such as \cite{zhou2017unsupervised}, without any labeled training data. By utilizing the ability of neural networks to learn a prior for the scale of a scene, these networks are able to predict accurate 6-dof poses and depths without the scale ambiguity problem that optimization based structure from motion methods face for monocular inputs, and without the need for explicit feature extraction. However, these methods abstract away the function between the image and the pose within the network weights, and so cannot provide any guarantees on robustness or safety. Furthermore, outliers such as independently moving objects remain a challenging problem, as incorporating outlier rejection schemes into the pose when it is directly regressed from the network has proven difficult.  

Most state-of-the-art learning frameworks for pose estimation have a network structure which estimates 6-dof pose from 2D images using convolutions. It is well established that CNN's are exceptionally powerful when operating on spatially structured data such as 2D images, and we have seen highly successful examples in problems such as depth \cite{godard2017unsupervised} and flow \cite{sun2018pwc} prediction, where network based methods are now state of the art. However, whether a network with these architectures can extract 3D translations and rotations from the 2D image plane is not obvious, and learning based methods for camera pose have yet to beat traditional methods. 

On the other hand, geometric optimization methods have seen immense success in this field with feature correspondences and outlier rejection using epipolar constraints~\cite{delmerico2018benchmark}. 
Given accurate correspondences between images and depths, optimization methods such as Random Sample Concensus (RANSAC)~\cite{fischler1981random} are able to recover extremely robust and accurate estimates of the camera pose and the surrounding scene. 

In this work, we decompose the classical direct visual odometry pipeline and introduce a learning-based front end. We combine the best of both worlds by leveraging the learning ability of neural networks to predict optical flow correspondences and disparities from a set of images, and applying a robust optimization backend using RANSAC to estimate relative pose from the network outputs. 

By applying RANSAC for pose estimation, we are able to apply outlier rejection at a pixel level to extract only the set of predicted flow and disparity values that best fit the camera pose model. In particular, most structure from motion (SFM) methods assume a static scene, and so are easily corrupted by independently moving objects. There have been a number of works that try to filter out these objects, for example by directly predicting a mask for the valid pixels in the scene from the network \cite{zhou2017unsupervised, vijayanarasimhan2017sfm}, or by detecting mismatches between the predicted optical flow and the disparity and pose \cite{ranjan2018adversarial, yang2018every}. Our method does this through a principled geometric matching, without the need to separately learn these objects. 

Given the pose estimated from RANSAC, we propose a set of additional refinement losses to further improve the performance of the proposed pipeline. Similar to previous works, we use a rigid motion model to estimate optical flow from the generated disparities and pose~\citep{zhou2017unsupervised}. However, we also perform the inverse, and estimate disparity from optical flow and pose. These new estimated values are used to apply additional photometric losses on the original network predictions, and we show that applying these losses at training time produces meaningful improvements in network performance.

Our contributions can be summarized as:
\begin{itemize}
    \item A novel end-to-end unsupervised structure from motion pipeline which uses two fully convolutional networks to predict optical flow and disparity from a pair of images, and a RANSAC outlier rejection scheme to robustly estimate instantaneous camera pose.
    \item A novel set of photometric losses that uses the camera pose to estimate flow from disparity and vice versa, allowing for additional supervision of each modality from another image.
    \item Evaluation on the KITTI datasets, where we show that our method is able to achieve state of the art performance with a very small network, as well as robustness to independently moving objects, with ablations demonstrating the improvements of our RANSAC method.
    
\end{itemize}
\section{Related Work}
There have been a number of recent methods that leverage the principles of photoconsistency and image warping \cite{jaderberg2015spatial} to perform unsupervised learning of image motion. Garg et al.~\cite{garg2016unsupervised} and Godard et al.~\cite{godard2017unsupervised} showed that metric depth can be learned from a monocular camera by warping stereo images onto one another. Similarly, Yu et al.~\cite{jason2016back} and Meister et al.~\cite{meister2017unflow} use a similar transformation to learn optical flow from a pair of images. Zhou et al.~\cite{zhou2017unsupervised} also showed that 3D camera motion can also be learned from this regime, by jointly predicting the egomotion of the camera, the depth of the scene, and combining the two to warp each pixel in the image. Since then, a number of methods have improved upon this scheme. Zhan et al~\cite{zhan2018unsupervised} add a feature reconstruction loss to resolve ambiguities seen with a photometric loss, such as in textureless regions. Li et al.~\cite{li2018undeepvo} use the egomotion to align the point clouds associated with the predicted depths, while Wang et al.~\cite{wang2018recurrent} introduce a recurrent neural network to replace the feed-forward CNN.

However, these methods rely on a rigid scene assumption, and so are corrupted by independently moving objects in the scene. A number of works have tried to remove these objects from the loss function, including the works by Zhou et al.~\cite{zhou2017unsupervised} and Vijayanarasimhan et al.~\cite{vijayanarasimhan2017sfm}, which predict masks to remove invalid pixels. Moreover, the works by Ranjan et al.~\cite{ranjan2018adversarial}, Luo et al.~\cite{luo2018every} and Yang et al.~\cite{yang2018every} use the mismatches between egomotion and depth and optical flow predictions to generate these masks. 

In a similar vein to our work, Wang et al.~\cite{Wang_2018_CVPR} train a depth network by applying direct visual odometry optimization using the predicted depths to minimize the image reprojection error, and Yang et al.~\cite{yang2018deep} incorporate a separately trained depth network into the Direct Sparse Odometry~\cite{engel2018direct} framework, providing a monocular odometry pipeline that is able to estimate metric trajectories comparable to a stereo setup.

Our work, on the other hand, uses an interpretable geometric, model-based backend to jointly detect outliers and regress pose, while leveraging the strength of the network in 2D to predict both correspondences and depths.
\section{Method}
Our pipeline consists of two neural networks, one which predicts optical flow from a pair of images, and one which predicts disparity from a single image. We have significantly reduced the size of the network, which we discuss in Sec.~\ref{sec:architecture}.

Each network can be trained separately, with a combination of photometric, smoothness and geometric losses, which we describe in Sec.~\ref{sec:appearance}-\ref{sec:total}. At training time, we sample four images randomly from the dataset, consisting of two stereo pairs from times $t_0$ and $t_1$. We then pass these images into both networks to predict a forward, $F_{t_0\rightarrow t_1}(\vec{x})$, and backward, $F_{t_1\rightarrow t_0}(\vec{x})$, flow for each image in the stereo pairs, and a disparity for each image, $D(\vec{x})$, resulting in four flow and disparity predictions respectively for every two stereo image pairs.

As both flow and disparity can be modeled by a general image warping function that warps a pixel, $\vec{x}$, to another location: $W(\vec{x}_i)\rightarrow \vec{x}_j$, we will describe our loss functions in terms of this general warping, which can then be substituted by either optical flow or disparity.

We use RANSAC~\citep{fischler1981random}, Sec.~\ref{sec:ransac}, for relative pose estimation and outlier rejection, taking the flow and disparity from the networks as inputs. 

The estimated pose is then used to apply a second set of \textit{refinement} losses Sec.~\ref{sec:refinement}. First, a rigid motion model is applied to estimate disparity from pose and flow, and flow from disparity and pose. Photometric losses are then applied to the spatial and temporal images warped by this second set of flow and disparity estimates.

This pipeline is depicted in Fig.~\ref{fig:pipeline}.

\subsection{Appearance Loss}
\label{sec:appearance}
Given a pair of images, $I_i$, $I_j$, and a warping between them, $W_{i\rightarrow j}$, we apply a photoconsistency assumption, which assumes that the correct warp should warp pixels from $I_i$ to pixels at $I_j$ with the same intensity. We therefore apply the following photometric loss to enforce the constraint:
\begin{align}
\mathcal{L}_{\text{photo}}^{W_{i\rightarrow j}}(\vec{x})=&\rho(I_i(\vec{x}) - I_j(W_{i\rightarrow j}(\vec{x})))\label{eq:appearance_loss}\\
\rho(x)=&\sqrt{x^2+\epsilon^2}
\end{align}
where $\rho(x)$ is the robust Charbonnier loss function used in \citep{jason2016back}.

We combine this photometric loss with a structural similarity (SSIM) loss \cite{wang2004image} to form our appearance loss:
\begin{align}
\mathcal{L}_{\text{appearance}}^{W_{i\rightarrow j}}(\vec{x})=& (1-\alpha)\text{SSIM}(\vec{x}) + \alpha \mathcal{L}_{\text{photo}}(\vec{x})
\end{align}
\subsection{Geometric Consistency}
Several works, such as \cite{godard2017unsupervised, meister2017unflow}, have proposed additional losses to constrain geometric consistency between the forward and backward (or left and right) estimates of a warp. That is, the backward warp, warped to the previous image, should be equivalent to the negative of the forward warp. We apply this constraint to both the disparity and flow:
\begin{align}
\mathcal{L}_{\text{consistency}}^{W_{i\rightarrow j}}(\vec{x})=&\rho(W_{i\rightarrow j}(\vec{x})+W_{j\rightarrow i}(W_{i\rightarrow j}(\vec{x})))
\end{align}

\subsection{Smoothness Regularization}
\label{sec:smoothness}
We further apply a constraint for the warp to be locally smooth. This is applied with an edge aware smoothness loss, which weighs the loss lower at pixels with high image gradient:
\begin{align}
\mathcal{L}_{\text{smooth}}^{W_{i\rightarrow j}}(\vec{x})=&\rho\left(\delta_x W_{i\rightarrow j}(\vec{x})e^{-|\delta_x I(\vec{x})|}\right) + \nonumber\\
&\rho\left(\delta_y W_{i\rightarrow j}(\vec{x})e^{-|\delta_y I(\vec{x})|}\right)
\end{align}

\subsection{Motion Occlusion Estimation}
\label{sec:occlusion}
For a given warp function, there may be pixels that are warped out of the image. Applying a photometric or geometric consistency loss at these pixels would introduce errors into the model, as we cannot sample from points from outside the image. To resolve this issue, we generate a mask, $M_{\text{occ}}^{W_{i\rightarrow}(\vec{x})}$, which is 0 for pixels that are warped out of the image, and 1 otherwise, and is used for the photometric and geometric consistency losses. Note that this mask is computed directly from the predicted flows and disparities, and does not contain any learnable components.

\subsection{Total Loss}
\label{sec:total}
For a single warp, $W_{i\rightarrow j}(\vec{x})$, the total loss is:
\begin{align}
\mathcal{L}^{W_{i\rightarrow j}}_{\text{total}}=&\sum_{\vec{x}} M_{occ}^{W_{i\rightarrow j}}(\vec{x})\left(\mathcal{L}_{\text{photo}}^{W_{i\rightarrow j}}(\vec{x}) + \lambda_1\mathcal{L}_{\text{consistency}}^{W_{i\rightarrow j}}(\vec{x})\right) \nonumber\\
&+ \lambda_2\mathcal{L}_{\text{smoothness}}^{W_{i\rightarrow j}}(\vec{x})
\end{align}
The final loss, then, is the sum of $\mathcal{L}_{\text{total}}$ for each of the flow and disparity predictions:
\begin{align}
\mathcal{L}_{\text{total}}=&\sum_{c\in\{l, r\}}\mathcal{L}_{\text{total}}^{F_{0\rightarrow 1, c}}+\mathcal{L}_{\text{total}}^{F_{1\rightarrow 0, c}} + \nonumber \\
&\sum_{t\in \{0, 1\}}\mathcal{L}_{\text{total}}^{D_{l\rightarrow r, t}} + \mathcal{L}_{\text{total}}^{D_{r\rightarrow l, t}}\label{eq:base_loss}
\end{align}
\subsection{RANSAC Outlier Rejection}
\label{sec:ransac}
Given the flow and disparity predictions, $F$, $D$, we can use the Random Sample Consensus (RANSAC) algorithm~\cite{fischler1981random} to estimate the best set of inliers to be used to estimate the pose of the cameras. RANSAC is a robust inlier selection scheme that allows us to filter out outliers such as independently moving objects from the optimization to estimate pose. The motion field equation that constrains pose, depth and optical flow is:
\begin{align}
&F(\vec{x})=\frac{1}{Z(\vec{x})}Av + B\omega \nonumber\\
&=\frac{1}{Z(\vec{x})}\begin{bmatrix}-1 & 0 & x \\ 0 & -1 & y\end{bmatrix}v+\begin{bmatrix} xy & -(1+x^2) & y \\ 1+y^2 & -xy & -x\end{bmatrix}\omega\label{eq:motion_field}\\
&Z(\vec{x})=\frac{fb}{D(\vec{x})}
\end{align}
where $x$ and $y$ represent the coordinates in the normalized camera frame of the corresponding pixel, $v$ and $\omega$ are the linear and angular velocity of the camera in the camera frame, $Z(\vec{x})$ is the depth in the camera frame, and $Av$ and $B\omega$ are the 2-dimensional vectors corresponding to linear and angular velocity terms. $f$ is the focal length of the camera and $b$ is the baseline between the cameras. Note that, in this work, we estimate displacements rather than velocity, although we will keep the same notation for brevity. i.e. $v\times t$ and $\omega\times t$ instead of $v$ and $\omega$.

Using $F(\vec{x})$ and $D(\vec{x})$ from the network, this is a least squares problem for $v$ and $\omega$. We perform 3-point RANSAC by randomly sampling 3 points from $F(\vec{x})$ and $D(\vec{x})$, and solving for $v$ and $\omega$. The solution is then used to find the set of inlier points that best fit the estimated model. The threshold for inliers is set to be the absolute difference between the network and RANSAC flow estimates, in terms of pixels. Because the motion field equation above assumes a static scene, the computed inlier mask will automatically filter out independently moving (non-static) objects.

\subsection{Refinement Loss}
\label{sec:refinement}
\begin{figure}
\centering
    \includegraphics[width=.4\linewidth]{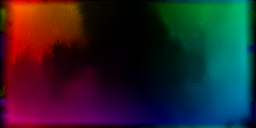}
    \includegraphics[width=.4\linewidth]{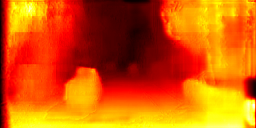}
    \includegraphics[width=.4\linewidth]{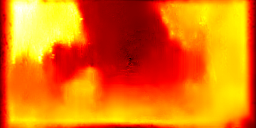}
    \includegraphics[width=.4\linewidth]{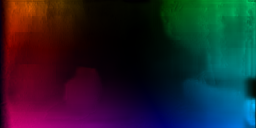}
\caption{Using the RANSAC pose, we can use the predicted flow (top left) to estimate disparity (bottom left), and disparity (top right) to estimate flow (bottom right). Best viewed in color.}
\label{fig:ransac_pred}
\end{figure}

Using the pose estimate from RANSAC and the motion field equation in \eqref{eq:motion_field}, we can estimate the optical flow from the disparities and pose using the forwards equation, as in Zhou et al.~\cite{zhou2017unsupervised}. However, as our flow predictions also directly contribute to the final pose estimate, we also estimate disparities using the flow and pose though the backwards equation derived from \eqref{eq:motion_field}. To estimate disparity from flow and pose, we would like to find the disparities, $\hat{D}(\vec{x}),$ that minimize $\|L(\vec{x})\|_2^2$, where:
\begin{align}
L(\vec{x}) =& F(\vec{x}) - \frac{\hat{D}(\vec{x})}{fb}Av - B\omega
\intertext{This is equivalent to finding the scale between the vector $v_1$ and the projection of $\vec{v_2}$ onto $\vec{v_1}$:}
\hat{D}(\vec{x})=& \frac{\vec{v_2}^T \vec{v_1}}{\vec{v_1}^T \vec{v_1}}\\
\vec{v_1} =&Av/fb , \vec{v}_2 = F(\vec{x}) - B\omega
\end{align}
Using these estimated flows and disparities, we can then compute the appearance loss, \eqref{eq:appearance_loss}, for the other image sequence. That is, we can use the disparities estimated from flow to compute the appearance loss on the stereo pair, and the flow estimates from disparity on the temporal pair. These losses allow each network to learn from an additional pair of images. An example of these estimates can be found in Fig.~\ref{fig:ransac_pred}

\subsection{Differentiability of RANSAC}
As noted in prior work by~\citet{brachmann2017dsac}, the hard thresholding to determine inliers and the argmax function used to compute the best set of inliers is non differentiable. However, given the set of inliers, the function to estimate camera velocity from the inlier set of flows and disparities is a simple matrix inversion, which is differentiable. In this work, we treat the inlier set as a fixed, non-learned, mask, similar to the motion occlusion mask in \ref{sec:occlusion}. As a result, gradient flows from the estimated pose through the inlier flows and disparities. This makes the refinement pipeline differentiable with respect to the set of inlier points, but not the selection of inliers themselves. 

Future work could involve testing this pipeline with a differentiable RANSAC function similar to DSAC~\cite{brachmann2017dsac}. However, this would require an additional CNN to score the quality of each velocity hypothesis to replace the hard thresholding for inliers, and increase the complexity of the system. This may only provide a marginal boost in performance for this relatively simple problem, when compared to traditional RANSAC which has shown great success in many odometry pipelines.
\begin{table*}[t]
\centering
 \begin{tabular}{ccccccccccccccc} 
 \toprule
  \textbf{Sequence} & \multicolumn{2}{c}{00} & \multicolumn{2}{c}{01} & \multicolumn{2}{c}{02} & \multicolumn{2}{c}{03} & \multicolumn{2}{c}{04} & \multicolumn{2}{c}{05} & \multicolumn{2}{c}{06}\\
  \hline \\[-2ex]
    & $t_{rel}$ & $r_{rel}$ & $t_{rel}$ & $r_{rel}$ & $t_{rel}$ & $r_{rel}$ & $t_{rel}$ & $r_{rel}$ & $t_{rel}$ & $r_{rel}$  & $t_{rel}$ & $r_{rel}$ & $t_{rel}$ & $r_{rel}$\\ \hline\\[-1ex]
    Ours & 4.95 & \textbf{1.39} & 45.5 & 1.78 & 6.40 & \textbf{1.92} & \textbf{4.83} & \textbf{2.11} & \textbf{2.43} & \textbf{1.16} & 3.97 & \textbf{1.20} & \textbf{3.49} & \textbf{1.02}\\
    SFMLearner~\cite{zhong2017self} & 66.4 & 6.13 & \textbf{35.2} & 2.74 & 58.8 & 3.58 & 10.8 & 3.92 & 4.49 & 5.24 & 18.7 & 4.10 & 25.9 & 4.80\\
    UnDeepVO~\cite{li2018undeepvo} & \textbf{4.14} & 1.92 & 69.1 & \textbf{1.60} & \textbf{5.58} & 2.44 & 5.00 & 6.17 & 4.49 & 2.13 & \textbf{3.40} & 1.50 & 6.20 & 1.98\\
    \citet{zhan2018unsupervised} & - & - & - & - & - & - & - & - & - & - & - & - & - & -\\
    \citet{luo2018every} & - & - & - & - & - & - & - & - & - & - & - & - & - & -\\
    \hline\\[-2ex]
    DVSO~\cite{yang2018deep} & 0.71 & 0.24 & 1.18 & 0.11 & 0.84 & 0.22 & 0.77 & 0.18 & 0.35 & 0.06 & 0.58 & 0.22 & 0.71 & 0.20\\
  \hline\\[-1ex]
  & \multicolumn{2}{c}{07} & \multicolumn{2}{c}{08} & \multicolumn{2}{c}{09*} & \multicolumn{2}{c}{10*} & \multicolumn{2}{c}{Mean*} & \multicolumn{2}{c}{Mean (all)}\\
 \hline \\[-2ex]
    & $t_{rel}$ & $r_{rel}$ & $t_{rel}$ & $r_{rel}$ & $t_{rel}$ & $r_{rel}$ & $t_{rel}$ & $r_{rel}$ & $t_{rel}$ & $r_{rel}$  & $t_{rel}$ & $r_{rel}$ \\ [0.5ex]
	\hline\\[-1ex]
    Ours & 4.50 & \textbf{1.78} & 4.08 & \textbf{1.17} & 4.66 & 1.69 & 6.30 & \textbf{1.59} & 5.48 & \textbf{1.64} & \textbf{8.28} & \textbf{1.59}\\
    SFMLearner~\cite{zhong2017self} & 21.3 & 6.65 & 21.9 & 2.91 & 18.8 & 3.21 & 14.3 & 3.30 & 16.6 & 3.26 & 27.0 & 4.23\\
    UnDeepVO~\cite{li2018undeepvo} & \textbf{3.15} & 2.48 & \textbf{4.08} & 1.79 & 7.01 & 3.61 & 10.6 & 4.65 & 8.82 & 4.13 & 11.2 & 2.75\\
    \citet{zhan2018unsupervised} & - & - & - & - & 11.9 & 3.60 &  12.6 & 3.43 & 12.3 & 3.52 & - & -\\
    \citet{luo2018every} & - & - & - & - & \textbf{3.72} & \textbf{1.60} & \textbf{6.06} & 2.22 & \textbf{4.89} & 1.91 & - & -\\
    \hline\\[-2ex]
    DVSO~\cite{yang2018deep}  & 0.73 & 0.35 & 1.03 & 0.25 & 0.83 & 0.21 & 0.74 & 0.21 & 0.79 & 0.21 & 0.77 & 0.20\\
 \bottomrule \\
\end{tabular}
\caption{Evaluation against competing instantaneous unsupervised pose learning methods, as well as DVSO~\citep{yang2018deep}, which uses a windowed optimization. Translation $t_{rel}$(\%) and rotation $r_{rel}$($^{\circ}/\SI{100}{m}$) RMSE on the 11 KITTI Odometry training sequences is presented. Bold indicates best instantaneous result for each column. Final means are computed over the training set (sequences 09 and 10) and all sequences, respectively. As SFMLearner~\citep{zhou2017unsupervised} and \citet{zhan2018unsupervised} are monocular methods, their results have been corrected for scale. Only results for 09 and 10 are provided in \citet{zhan2018unsupervised} and \citet{luo2018every}.} 
\label{tab:pose_eval}
\end{table*}
\begin{table*}[t]
\centering
 \begin{tabular}{ccccccccccccc} 
 \toprule
  \textbf{Sequence} & \multicolumn{2}{c}{00} & \multicolumn{2}{c}{01} & \multicolumn{2}{c}{02} & \multicolumn{2}{c}{03} & \multicolumn{2}{c}{04} & \multicolumn{2}{c}{05}\\
  \hline \\[-2ex]
    & $t_{rel}$ & $r_{rel}$ & $t_{rel}$ & $r_{rel}$ & $t_{rel}$ & $r_{rel}$ & $t_{rel}$ & $r_{rel}$ & $t_{rel}$ & $r_{rel}$  & $t_{rel}$ & $r_{rel}$\\ [0.5ex]
	Ours & \textbf{4.95} & \textbf{1.39} & \textbf{45.5} & 1.78 & 6.40 & 1.92 & 4.83 & 2.11 & 2.43 & 1.16 & 3.97 & \textbf{1.20}\\
    Ours (no refinement) & 5.45 & 1.78 & 63.9 & 1.53 & \textbf{6.21} & \textbf{1.76} & \textbf{4.27} & \textbf{1.79} & \textbf{2.05} & \textbf{1.04} & \textbf{3.82} & 1.38\\
  \hline\\[-1ex]
  & \multicolumn{2}{c}{06} & \multicolumn{2}{c}{07} & \multicolumn{2}{c}{08} & \multicolumn{2}{c}{09} & \multicolumn{2}{c}{10} & \multicolumn{2}{c}{Mean}\\
 \hline \\[-2ex]
    & $t_{rel}$ & $r_{rel}$ & $t_{rel}$ & $r_{rel}$ & $t_{rel}$ & $r_{rel}$ & $t_{rel}$ & $r_{rel}$ & $t_{rel}$ & $r_{rel}$  & $t_{rel}$ & $r_{rel}$ \\ [0.5ex]
   	Ours & \textbf{3.49} & \textbf{1.02} & \textbf{4.50} & \textbf{1.78} & 4.08 & 1.17 & \textbf{4.66} & \textbf{1.69} & \textbf{6.30} & \textbf{1.59} & \textbf{8.28} & \textbf{1.59}\\
   	Ours (no refinement) & 4.02 & 1.63 & 5.47 & 2.56 & \textbf{3.73} & \textbf{1.09} & 5.11 & 1.95 & 6.71 & 2.52 & 10.06 & 1.73\\
 \bottomrule \\
\end{tabular}
\caption{Pose ablation study on the effect of the proposed refinement losses with models trained with and without the proposed refinement losses. Models were trained on sequences 00-08.}
\label{tab:pose-ablation}
\end{table*}
\begin{figure*}
\centering
    \includegraphics[width=.25\linewidth]{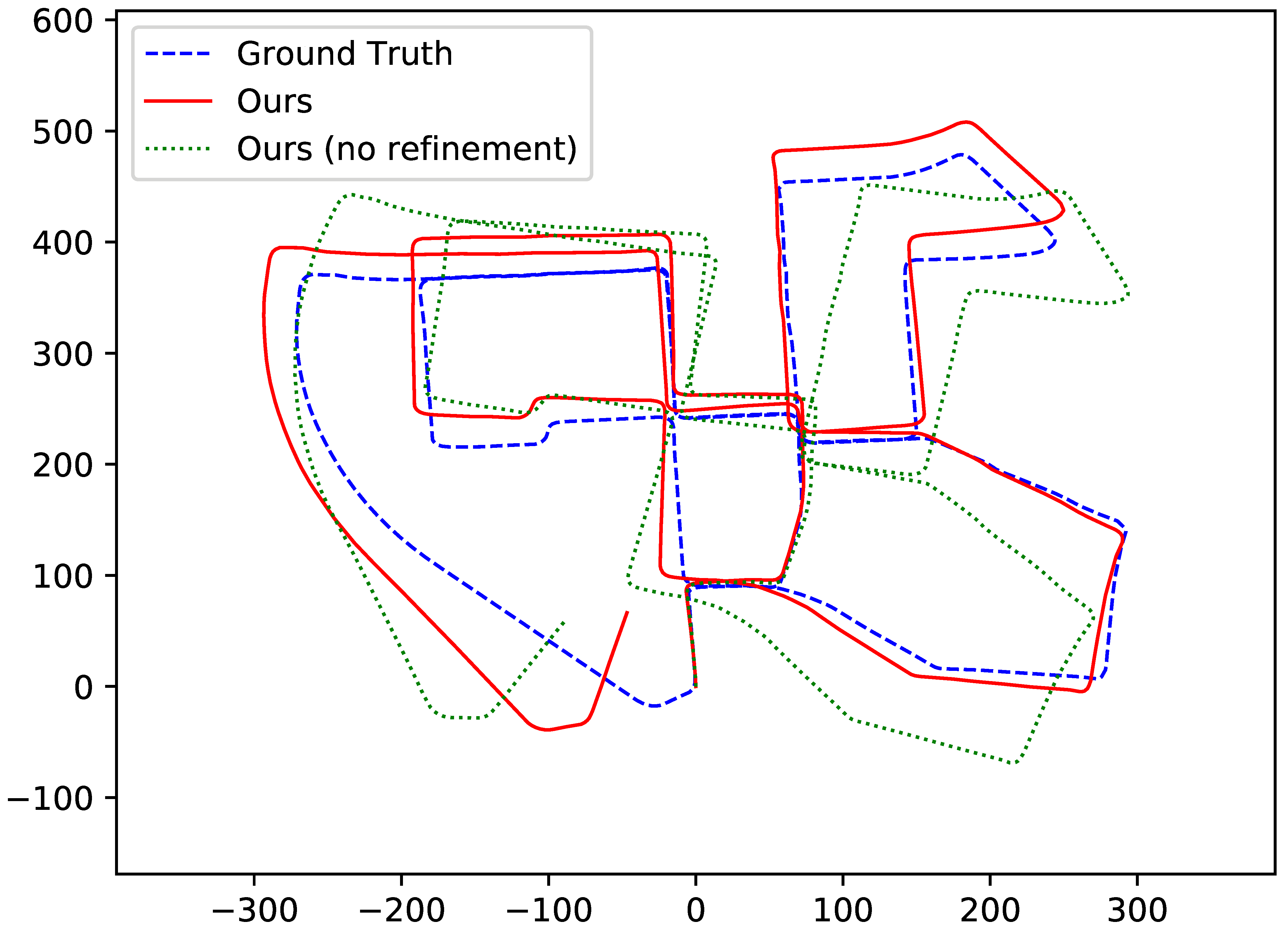}
    \includegraphics[width=.25\linewidth]{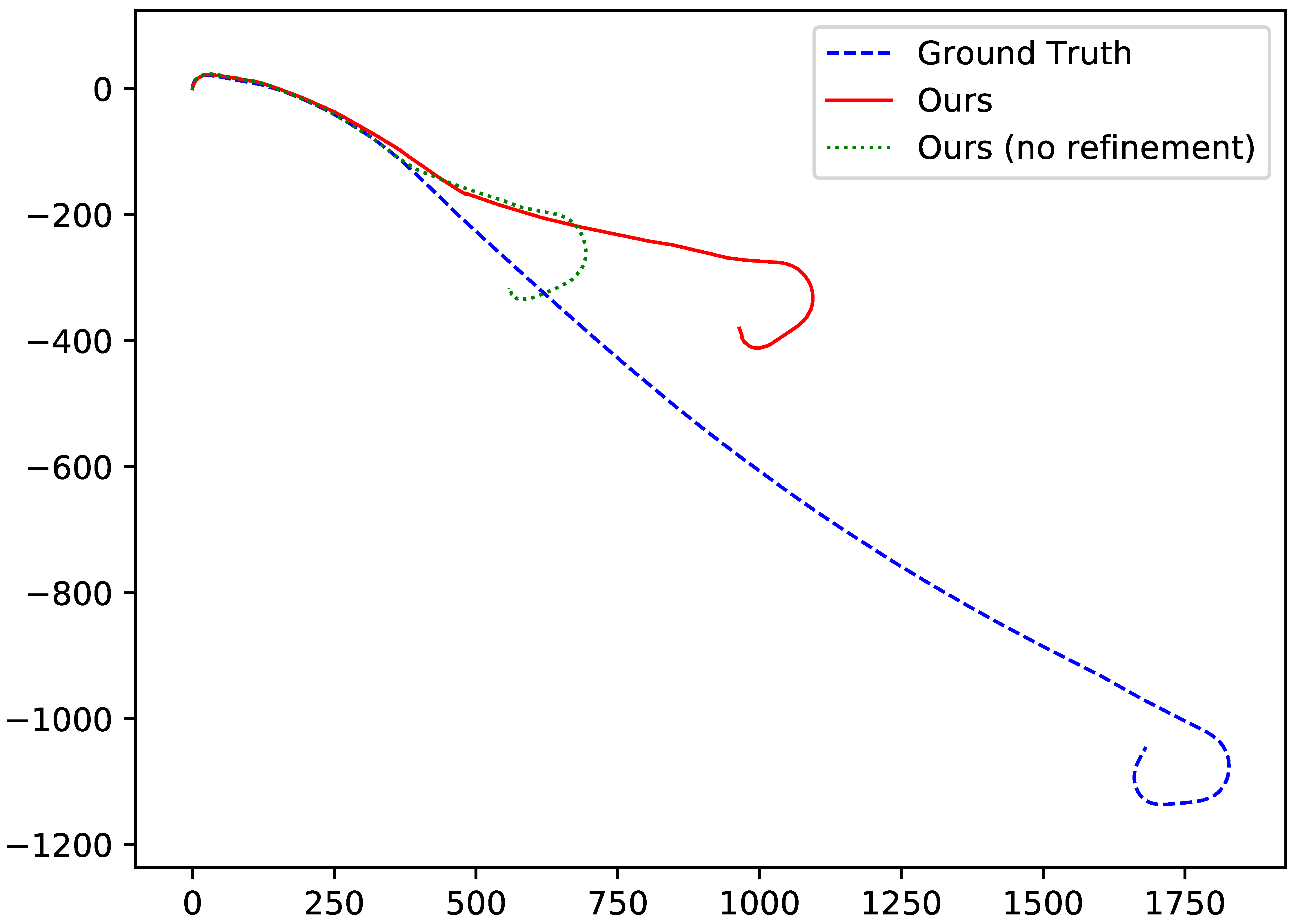}
    \includegraphics[width=.25\linewidth]{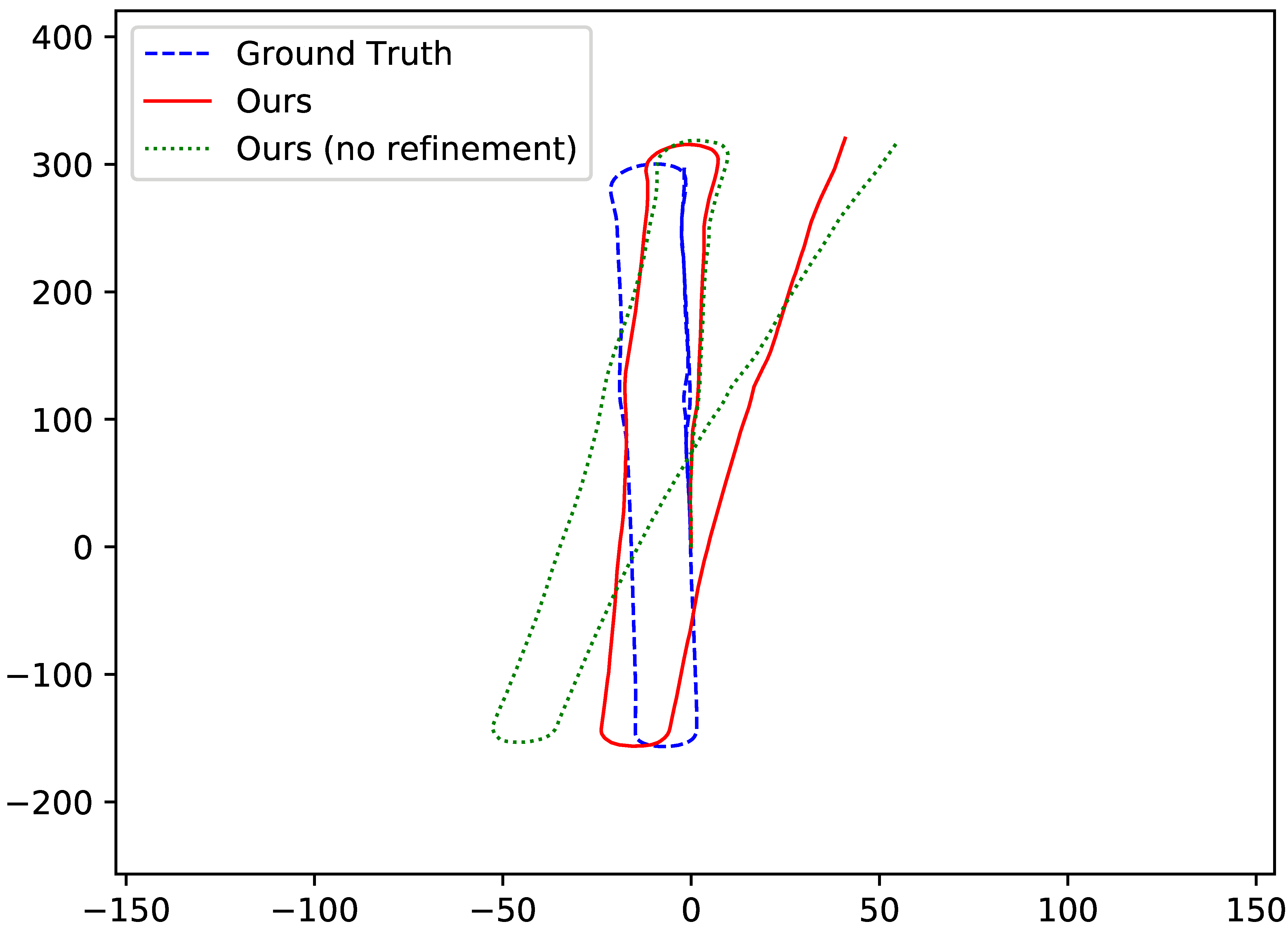}
    \includegraphics[width=.25\linewidth]{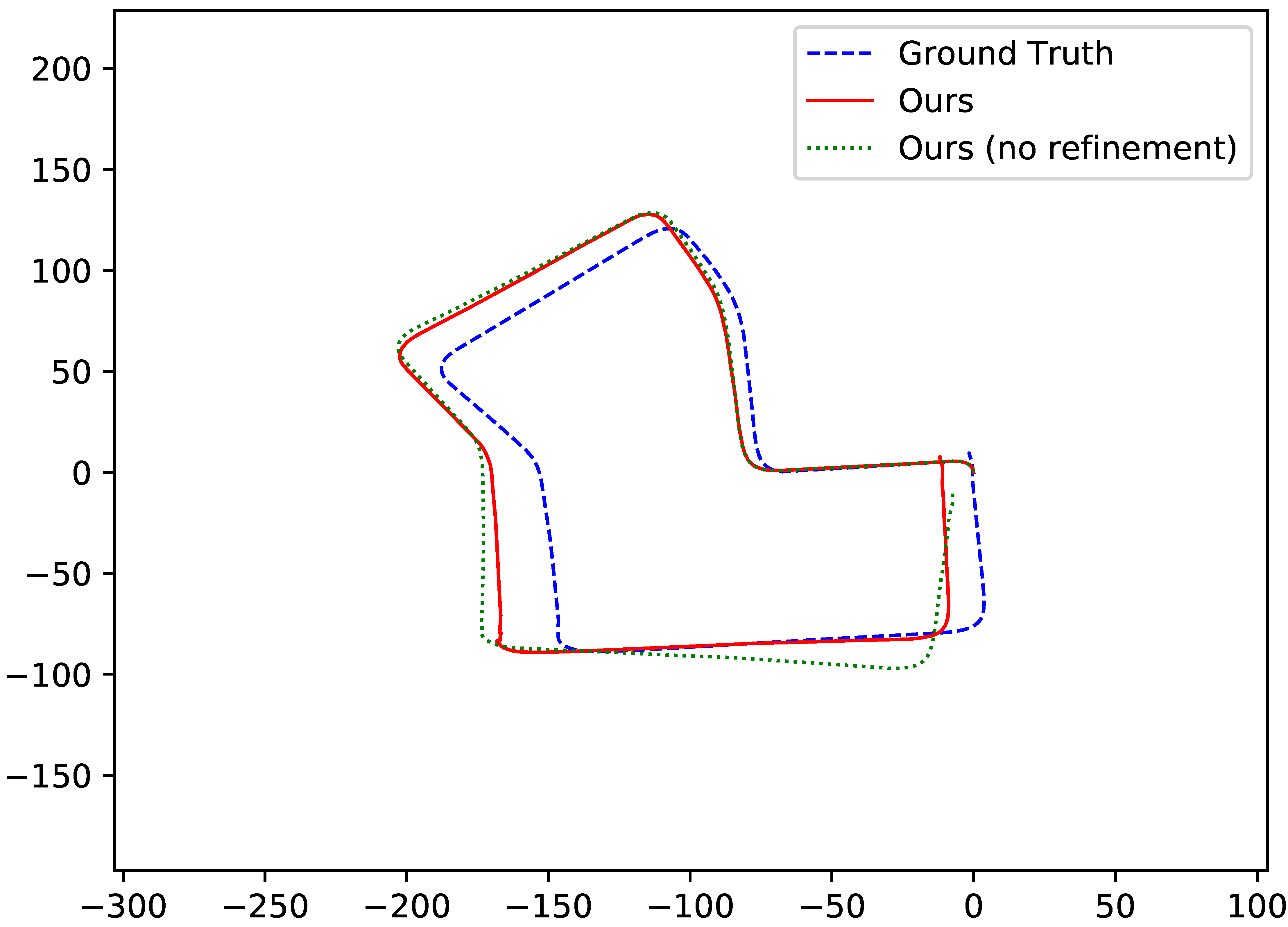}
    \includegraphics[width=.25\linewidth]{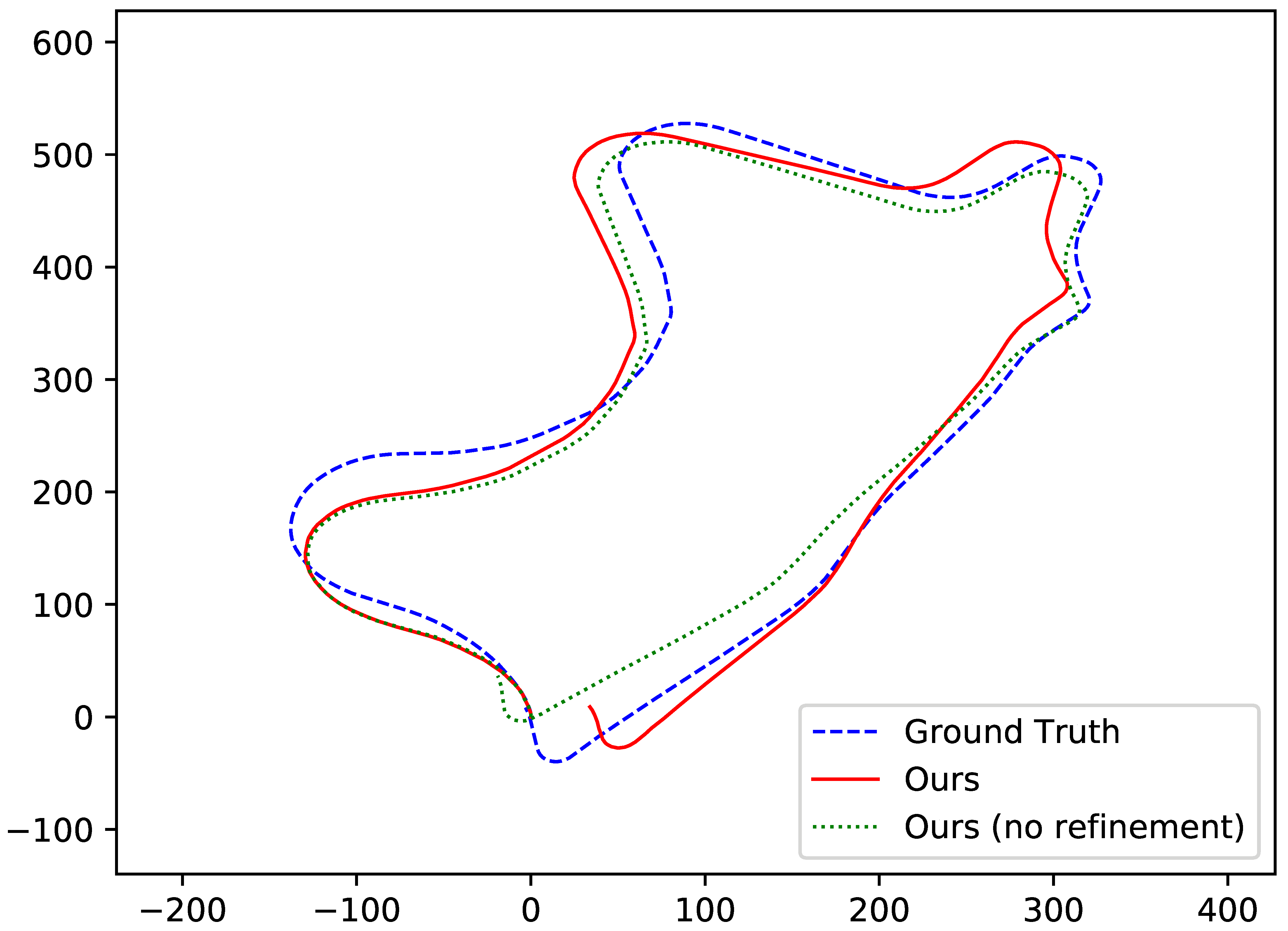}
    \includegraphics[width=.25\linewidth]{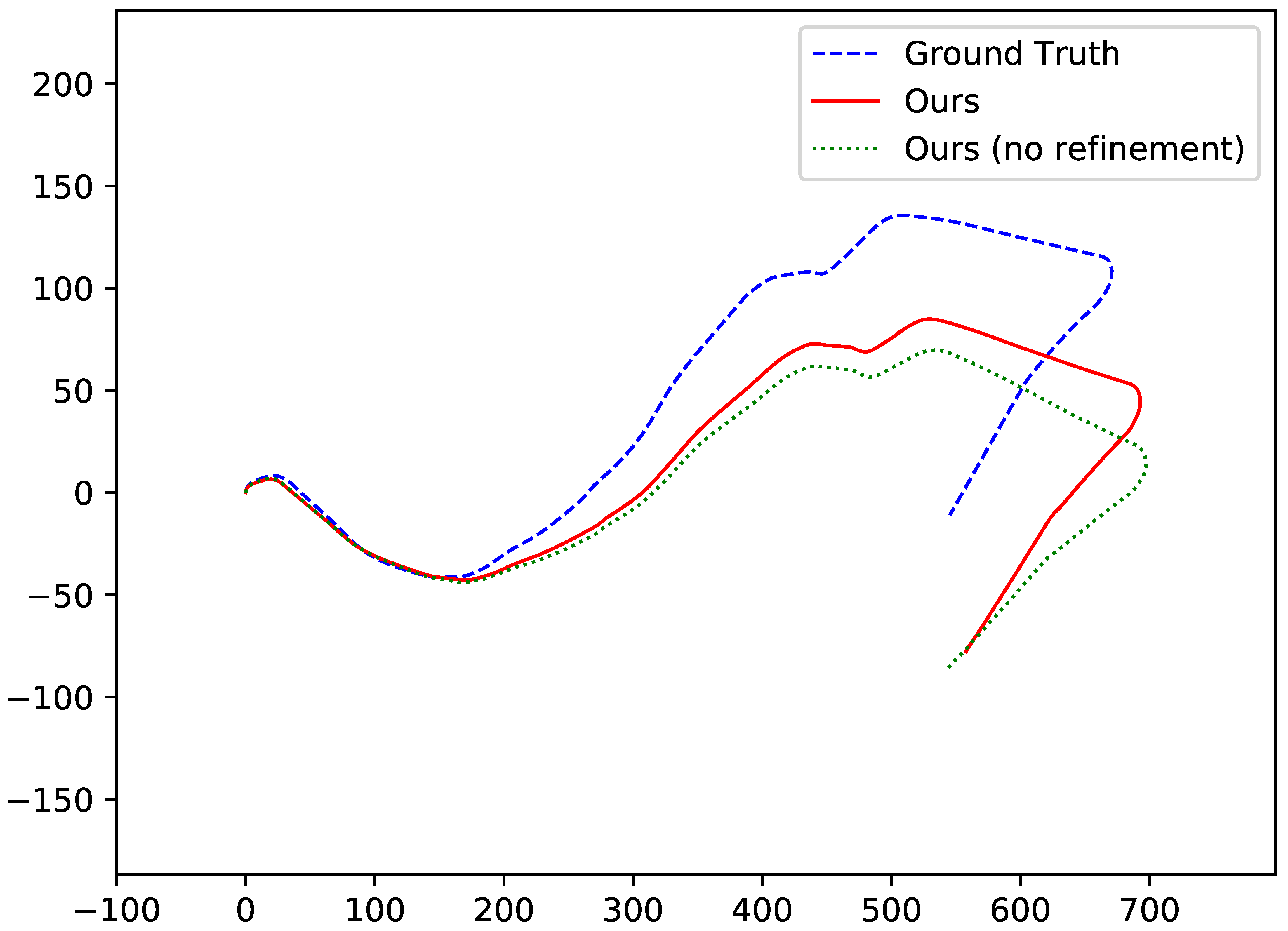}
\caption{Selected trajectories of the models trained on KITTI raw before and after refinement, as well as our final model, against ground truth. Sequence numbers from top left to bottom right are 00, 01, 06, 07, 09, 10. Best viewed in color.}
\label{fig:trajectories}
\end{figure*}
\subsection{Network Architecture}
\label{sec:architecture}
One disadvantage of our method is that it requires dense per-pixel optical flow and disparity predictions from two networks in order to estimate pose, as compared to other networks that directly regress pose using a, potentially smaller, CNN. In order to compensate for this, we use a much smaller pair of networks in our pipeline than prior works, such as \cite{godard2017unsupervised, meister2017unflow}. Our network is based on the Flownet-S architecture~\cite{fischer2015flownet}, but with a significantly smaller model. In short, we remove the final three convolution layers from the encoder and the first two convolution layers from the decoder, and halve the number of output channels at each layer. In our experiments, our network reduces an input image with resolution 128x448 to activations with resolution 8x28 and 256 channels at the bottleneck. 

In Tab. \ref{tab:network_size}, we compare the number of parameters in this proposed architecture with several methods, including Monodepth~\cite{godard2017unsupervised}, SFM-Learner~\cite{zhou2017unsupervised}, FlowNet~\cite{fischler1981random}, as well as the Resnet50~\cite{he2016deep} variant from Monodepth. Our mini network is less than 10\% of the size of these networks.

\begin{table}[t]
\begin{center}
\begin{tabular}{ c|c }
\hline\\ 
 Method & \# Parameters\\ \hline \\[-2ex]
 Ours & 2,943,496\\
 Monodepth & 31,596,936 \\ 
 SFM-Learner & 33,187,568\\
 FlowNet-S & 35,885,068\\
 Monodepth Resnet50 & 58,445,736\\
 \hline
\end{tabular}
\caption{Comparison of model size.}
\label{tab:network_size}
\end{center}
\end{table}

\begin{table*}[t!]
\centering
\begin{tabular}{c|c|cccc|ccc} 
\toprule
    & Split & RMSE & RMSE(log) & Abs Rel & Sq Rel & $\delta < 1.25$ & $\delta<1.25^2$ & $\delta<1.25^3$\\
  \hline \\[-1ex]
    Ours (no refinement) & stereo2015 & 6.753 & 0.274 & 0.182 & 2.570 & 0.807 & 0.924 & 0.963\\
    Ours & stereo2015 & 6.394 & 0.257 & 0.168 & 2.114 & 0.820 & 0.931 & 0.968 \\
    \hline\\[-2ex]
    \hline\\[-1ex]
    Ours (no refinement) & eigen & 4.240 & 0.218 & 0.147 & 0.897 & 0.816 & 0.938 & 0.975\\
    Ours & eigen & 4.366 & 0.230 & 0.154 & 1.021 & 0.808 & 0.933 & 0.972\\
    \hline\\[-1ex]
    DVSO~\cite{yang2018deep} & eigen & \textbf{3.390} & \textbf{0.177} & \textbf{0.092} & \textbf{0.547} & \textbf{0.898} & \textbf{0.962} & \textbf{0.982} \\
    \citet{godard2017unsupervised} resnet pp & eigen & 3.729 & 0.194 & 0.108 & 0.657 & 0.873 & 0.954 & 0.979\\
    \citet{garg2016unsupervised} & eigen & 5.104& 0.273& 0.169& 1.080& 0.740& 0.904& 0.962\\
    SFMLearner~\cite{zhong2017self} & eigen & 5.181 & 0.264 & 0.201 & 1.391 & 0.696 & 0.900 & 0.966\\
    UnDeepVO~\cite{li2018undeepvo} & eigen &  6.570 & 0.268 & 0.183 & 1.730 & - & - & -\\
    \citet{zhan2018unsupervised} & eigen & 4.204 & 0.216 & 0.128 & 0.815 & 0.835 & 0.941 & 0.975\\
\bottomrule 
\end{tabular}
\caption{Depth evaluation results on the Eigen (\SI{50}{m} cap) test split and KITTI stereo 2015.  Left: lower is better, right: higher is better.}
\label{tab:depth}
\end{table*}
\section{Experiments}
We evaluate our proposed pipeline on the KITTI Dataset~\cite{geiger2013vision} and make comparisons to other state-of-the-art networks for depth, flow and pose estimation. In addition, we provide ablation studies to evaluate the contribution of the proposed refinement losses in Sec.~\ref{sec:refinement}, as well as the effect of model size described in Sec.~\ref{sec:architecture} on performance. We also show qualitative examples where the outlier mask is able to correctly segment independently moving objects as outliers in Fig.~\ref{fig:qualitative}.
\subsection{Implementation Details}
\label{sec:implementation}
Our model was trained by first training each network for 8 epochs without the refinement losses, and then training with refinement losses for a total of 50 epochs. Input images were resized using bilinear interpolation to 128x448 pixels for both models. 100 RANSAC steps were run at each step, with an estimate considered an inlier if the norm between the RANSAC flow and network flow was less than 1 pixel.

Models used for pose evaluation were trained on the KITTI Odometry training sequences 00-08. Models used for optical flow and depth evaluation were trained on a subset of the KITTI Raw dataset, consisting of the city, residential and road sequences, with all images overlapping with the Eigen depth test split~\cite{eigen2014depth}, KITTI Stereo 2015 dataset and the KITTI Flow 2012 and 2015 datasets removed. Note that this is strictly less training data than training with each individual train/test split.
\subsection{Results}
We perform evaluations of each component in our pipeline on the KITTI Odometry 2012 and Flow and Stereo 2012 and 2015 datasets. 

\subsubsection{Pose Results}
For quantitative pose evaluation, we compute average RMSE translation (\%) and rotation (${}^\circ / 100m$) drift for all methods, as prescribed by \citet{Geiger2012CVPR}. We compare our method against the ablation studies, as well as competing unsupervised deep SFM methods SFMLearner~\citep{zhou2017unsupervised}, UnDeepVO~\cite{li2018undeepvo}, \citet{zhan2018unsupervised}, \citet{luo2018every} and DVSO~\cite{yang2018deep}. In addition, we provide comparisons against the monocular optimization based visual odometry pipeline ORBSLAM~\citep{murTRO2015}. As SFMLearner and Luo et al. are monocular methods that generate poses up to a scale, the trajectories are scale aligned before computing each error. These results can be found in Tab.~\ref{tab:pose_eval}.

From these results, our method outperforms UnDeepVO and Zhan et al., the instantaneous methods that predict up to scale depths across almost all sequences. In addition, these results strongly outperform monocular optimization frameworks which are unable to resolve the scale ambiguity such as ORBSLAM~\citep{murTRO2015}, as noted in the results in UnDeepVO. We also outperform SFMLearner, and perform comparably to Luo et al. on the test set. However, as Luo et al. only regress depths up to a scale factor, the pose results have been compensated for this scale. Overall, DVSO strongly outperforms all instantaneous methods. However, this is to be expected as DVSO incorporates a windowed optimization method to reduce drift, which is the main contributer of error in instantaneous methods. We present our method as a step towards closing the gap between instantaneous and more global methods, with the additional benefit of providing interpretable pose results compared to learning only methods.

\begin{figure}
\centering
\includegraphics[width=1.0\linewidth]{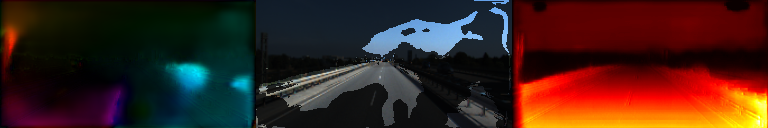}
\caption{An example of a failure case from sequence 01. From left to right are flow, inlier mask and disparity respectively. The network consistently underestimates the flow over the textureless road section, causing RANSAC to select a large part of the sky as inliers. Best viewed in color.}
\label{fig:failure}
\end{figure}

For qualitative results, Fig.~\ref{fig:trajectories} shows trajectories generated by the two models in our ablation against ground truth on 6 of the 11 sequences. Note that sequence 01 has a much higher translation error than the other datasets. This is because this is the only sequence on a highway in a very open space without many other structures, as can be seen in Fig.~\ref{fig:failure}. The network consistently underestimates the flow over the textureless road section, causing RANSAC to select a large part of the sky in the inlier set, and produce an underestimate of the translation. Note also that, despite the high translation error, the pipeline performs very well in terms of rotation error for sequence 01.

\subsubsection{Depth Results}
We evaluate our depth network prediction based on the test split and metrics used in Eigen et al.~\cite{eigen2014depth}. We compare our results according to the \SI{50}{m} cap. The Eigen test split has around 697 images in total selected from KITTI Raw sequences. In addition, we test our model on the KITTI Stereo 2015 dataset~\citep{Menze2018JPRS} using the same metrics. Both sets of experiments use the crop from \citet{garg2016unsupervised}. We report both results and compare to other methods, and report the values as in Table~\ref{tab:depth}. 

From these results, our model outperforms \citet{garg2016unsupervised}, SFMLearner~\citep{zhou2017unsupervised} and UnDeepVO~\citep{li2018undeepvo}, while performing comparably to \citet{zhan2018unsupervised}, which uses a larger image of 160x680 pixels, and performs worse than \citet{godard2017unsupervised} and DVSO~\citep{yang2018deep}, which use images of 256x512 pixels, and are semi-supervised in the case of DVSO, by depths initialized by DSO\citep{engel2018direct}.

\subsubsection{Flow Results}

Our optical flow evaluation is performed on KITTI Flow 2015 training dataset~\citep{Menze2015ISA}. We have the ground truth from 200 images in the training set and we calculate the average Endpoint Error (EPE) as an error metric. We then calculate the percentage of outliers (flows with EPE $>$ 3 and $>5\%$ of the true magnitude) on both the non-occluded (noc) pixels and the whole image (all) using the metrics defined by the KITTI Dataset~\cite{geiger2013vision}. 

We compare to other state-of-the-art unsupervised methods including the work by \citet{luo2018every}, UnFlow~\cite{meister2017unflow}, and GeoNet~\cite{yin2018geonet}. From these results, we perform significantly worse than the competing methods, likely due to the lower capacity and smaller input resolution in our network. However, we show in Sec.~\ref{sec:inlier_effect} that our RANSAC method is able to reject the high error flow values, resulting in statistics that outperform the competitors in these metrics.

\subsection{Ablation Study}
We perform an ablation study on the effect of the proposed refinement losses, by training an additional network without these losses (i.e. with \eqref{eq:base_loss} as the loss). Results for pose, depth and flow can be found in Tab.~\ref{tab:pose-ablation}, \ref{tab:depth}, \ref{tab:flow}, respectively.

From these results, we can see that the refinement losses result in improvements in pose and flow, with a neutral effect on depth. This corresponds from our observations during training that disparity estimation is an easier problem than optical flow, for a dataset with a consistent environment. Disparity estimation is a 1D search problem, and is constrained to be positive, while flow is 2D and can also be negative. By estimating disparity from flow, our refinement losses allow our network to reduce a component of the optical flow problem into a 1D search in the positive disparity space. On the other hand, estimating flow from disparity converts a 1D problem to 2D, and provides negligible gains. This improvement in flow error also helps to validate the accuracy of the estimated pose, as the refinement losses are only valid provided an accurate pose estimate.

\begin{table}
\centering
\begin{tabular}{ccc|cc} 
\toprule
    & \multicolumn{2}{c}{EPE} & \multicolumn{2}{c}{Error Perc}\\
	    & noc & all & noc & all\\
  \hline \\[-2ex]
    Ours & 10.66 & 19.74 & 45.84\% & 54.19\% \\
    Ours (no refinement) & 11.80 & 21.16 & 46.77\% & 54.99\% \\
    Luo et al.~\cite{luo2018every} & \textbf{3.86} & \textbf{5.66} & - & - \\
    UnFlow~\cite{meister2017unflow} & - & 8.10 & - & 23.27\%\\ 
    GeoNet~\cite{yin2018geonet} & 8.05 & 10.81 & - & -\\ 
\bottomrule
\end{tabular}
\caption{Comparison and ablation of flow prediction on the KITTI Flow 2015 dataset. noc = occluded pixels removed. A pixel is considered an outlier if its EPE is higher than 3 pixels and 5\% of its true magnitude.}
\label{tab:flow}
\end{table}
\subsection{Inlier Quality}
\label{sec:inlier_effect}
In this section, we investigate the quality of the flows and depths selected by RANSAC as inliers. In Tab.~\ref{tab:inlier_only_flow} and Tab.~\ref{tab:inlier_only_depth}, we present the flow and depth errors, computed only over inlier points. For the depth results, we also provide results without the crop from \citet{garg2016unsupervised}, as this already removed many difficult points that would be marked as outliers. From these results, we can also see that RANSAC selects points that have significantly lower error than the average. From the flow results, we can see that the ``noc" and ``all" results are very similar, suggesting that almost all of the occluded points have been rejected correctly by RANSAC. Essentially, by using RANSAC for inlier selection, we are able to estimate camera velocity from the equivalent of a much better model. 

This is likely possible due to the fact that the network tended to learn easier parts of the image first, such as high gradient corners, before slowly learning progressively harder regions such as textureless areas. Thus, the network generates reasonable flow and disparity values for at least some pixels, even early on in training or for very small networks. RANSAC is then able to select these correct values, even in the presence of challenging outliers.
\subsection{Independently Moving Objects}
\begin{figure*}
\centering
    \includegraphics[width=0.24\linewidth]{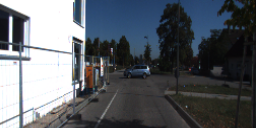}
    \includegraphics[width=0.24\linewidth]{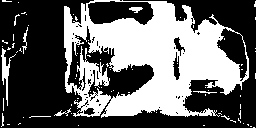}
    \includegraphics[width=0.24\linewidth]{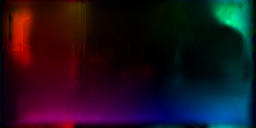}
    \includegraphics[width=0.24\linewidth]{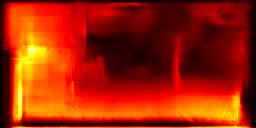}
    
    \includegraphics[width=0.24\linewidth]{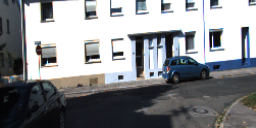}
    \includegraphics[width=0.24\linewidth]{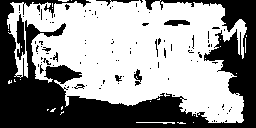}
    \includegraphics[width=0.24\linewidth]{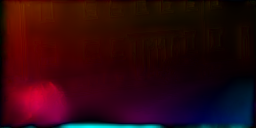}
    \includegraphics[width=0.24\linewidth]{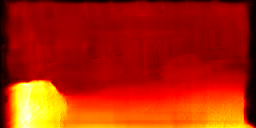}
    
    \includegraphics[width=0.24\linewidth]{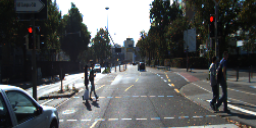}
    \includegraphics[width=0.24\linewidth]{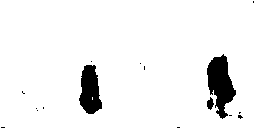}
    \includegraphics[width=0.24\linewidth]{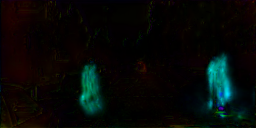}
    \includegraphics[width=0.24\linewidth]{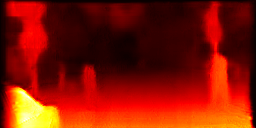}

    \includegraphics[width=0.24\linewidth]{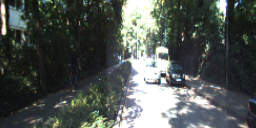}
    \includegraphics[width=0.24\linewidth]{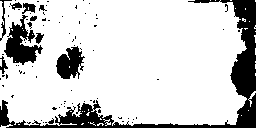}
    \includegraphics[width=0.24\linewidth]{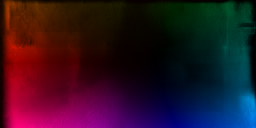}
    \includegraphics[width=0.24\linewidth]{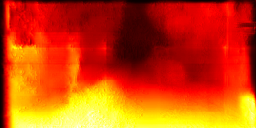}
    
    \includegraphics[width=0.24\linewidth]{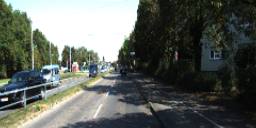}
    \includegraphics[width=0.24\linewidth]{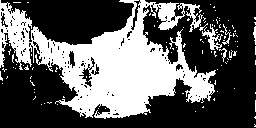}
    \includegraphics[width=0.24\linewidth]{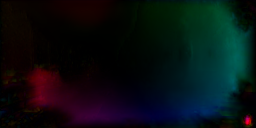}
    \includegraphics[width=0.24\linewidth]{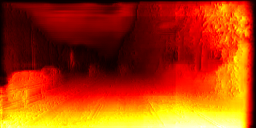}
    
    \includegraphics[width=0.24\linewidth]{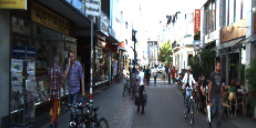}
    \includegraphics[width=0.24\linewidth]{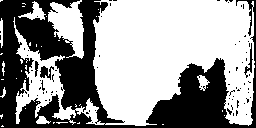}
    \includegraphics[width=0.24\linewidth]{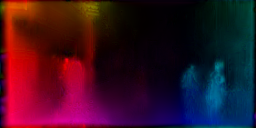}
    \includegraphics[width=0.24\linewidth]{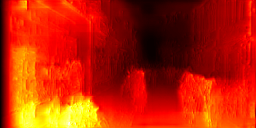}

    \includegraphics[width=0.24\linewidth]{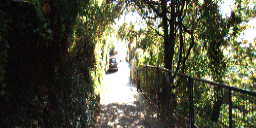}
    \includegraphics[width=0.24\linewidth]{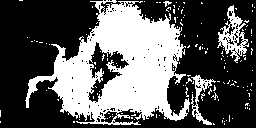}
    \includegraphics[width=0.24\linewidth]{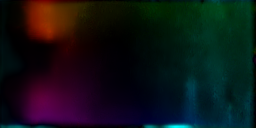}
    \includegraphics[width=0.24\linewidth]{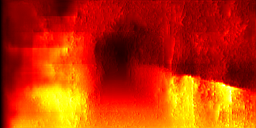}
    
        \includegraphics[width=0.24\linewidth]{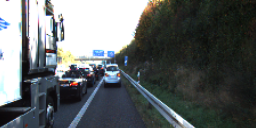}
    \includegraphics[width=0.24\linewidth]{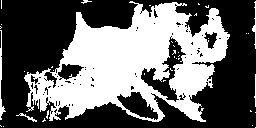}
    \includegraphics[width=0.24\linewidth]{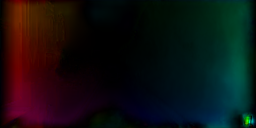}
    \includegraphics[width=0.24\linewidth]{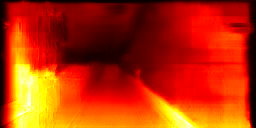}
    
        \includegraphics[width=0.24\linewidth]{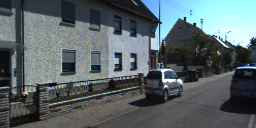}
    \includegraphics[width=0.24\linewidth]{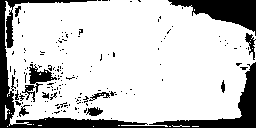}
    \includegraphics[width=0.24\linewidth]{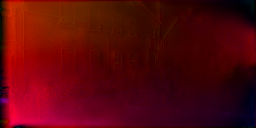}
    \includegraphics[width=0.24\linewidth]{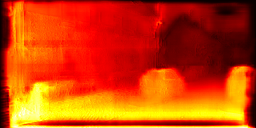}
    
        \includegraphics[width=0.24\linewidth]{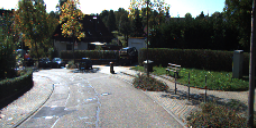}
    \includegraphics[width=0.24\linewidth]{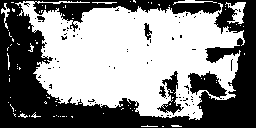}
    \includegraphics[width=0.24\linewidth]{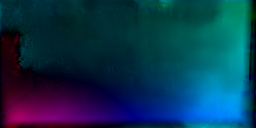}
    \includegraphics[width=0.24\linewidth]{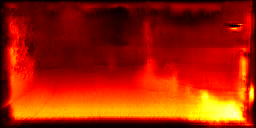}

\caption{Interesting qualitative results from our pipeline. Left to right: Grayscale image, inlier mask, predicted flow, predicted disparity. The inlier mask demonstrates RANSAC's ability to remove areas over which the network has high uncertainty, such as vegetation and textureless regions such as the sky. In addition, it allows us to automatically remove independently moving objects from the scene. Note that in the fourth image, there is a person slight left of the center of the image. Best viewed in color.}
\label{fig:qualitative}
\end{figure*}

We show some examples of RANSAC segmenting independently moving objects in Fig.~\ref{fig:qualitative}. As the outlier rejection is purely geometric, it does not rely on the network to learn that these objects have independent motion. The outlier masks also show that RANSAC segments regions where the networks tend to struggle, such as strong shadows (second row), sky (fifth row), thin structures (sixth row) and vegetation (seventh row).

\begin{table}[t]
\centering
\begin{tabular}{ccc|cc} 
\toprule
   & \multicolumn{2}{c}{EPE on Inliers} & \multicolumn{2}{c}{Error Perc. on Inliers}\\
    & noc & all & noc & all\\
  \hline \\[-2ex]
  Ours & \textbf{3.21} & \textbf{3.56} & \textbf{20.81} & \textbf{21.12} \\
  Ours (no refinement) & 3.25 & 3.39 & 21.07 & 21.24\\
\bottomrule
\end{tabular}
\caption{Evaluation of inlier quality with flow errors on the KITTI Flow 2015 dataset. Errors are computed only over inlier pixels, $\sim$21$\%$ of pixels.}
\label{tab:inlier_only_flow}
\end{table}

\begin{table}[t]
\centering
\begin{tabular}{ccccc} 
\toprule
    & RMSE & RMSE (log) & Abs Rel & Sq Rel\\
  \hline \\[-2ex]
    Ours & 12.63 & 0.30 & 0.24 & 22.63\\
    Ours (inliers) & \textbf{8.00} & \textbf{0.24} & \textbf{0.16} & \textbf{3.63}\\
    Ours (inliers, no ref.) & 9.00 & \textbf{0.24} & 0.17 & 7.92 \\
\bottomrule
\end{tabular}
\caption{Evaluation of inlier quality with depth errors from the KITTI Stereo 2015 dataset, without the crop from \citet{garg2016unsupervised}. The first row is computed over all pixels, while the last two are computed only over inlier pixels, $\sim$35$\%$ of pixels.}
\label{tab:inlier_only_depth}
\end{table}
\section{Conclusion}
In this work, we demonstrate a novel structure from motion pipeline that combines unsupervised learning with geometric optimization. Our method is able to compete with state of the art methods which directly predict pose from a network, while providing extra guarantees about robustness and automatic detection of independently moving objects. We show that this pipeline is able to achieve state of the art accuracy with a significantly reduced model, and hope that it can spur future work in bringing robustness and safety to learning for egomotion. We also show the contribution of our proposed refinement losses, as well as how the extracted inliers have similar error statistics to a much better set of models.

\section{Acknowledgements}
This work was supported in part by the Semiconductor Research Corporation (SRC) and DARPA. We also gratefully appreciate support through the following grants: NSF-DGE-0966142 (IGERT), NSF-IIP-1439681 (I/UCRC), NSF-IIS-1426840, NSF-IIS-1703319, NSF MRI 1626008, ARL RCTA W911NF-10-2-0016, ONR N00014-17-1-2093, the Honda Research Institute and the DARPA FLA program.

\bibliographystyle{plainnat}
\bibliography{ref}

\end{document}